\documentclass{article}

\usepackage[final]{neurips_2023}
\usepackage{amsmath}

\usepackage[utf8]{inputenc} 
\usepackage[T1]{fontenc}    
\usepackage{hyperref}       
\usepackage{url}            
\usepackage{booktabs}       
\usepackage{amsfonts}       
\usepackage{nicefrac}       
\usepackage{microtype}      
\usepackage{xcolor}         
\usepackage{graphicx}       
\usepackage{subfigure}
\usepackage{hyperref}
\usepackage{algorithm}
\usepackage{algorithmic}

\usepackage{amsmath}
\usepackage{diagbox}
\newcommand\numberthis{\addtocounter{equation}{1}\tag{\theequation}}

\title{Addressing Uncertainty in LLMs: Leveraging Semantic Entropy for Predicting Conformal Sets}
\title{Leveraging Semantic Clustering for Addressing Uncertainty in LLMs via UQ and Conformal Set Prediction}
\title{Enhancing Semantic Clustering for Uncertainty Quantification \& Conformal Prediction by LLMs}
\title{Addressing Uncertainty in LLMs: Enhancing Semantic Clustering for Uncertainty Quantification \& Conformal Prediction}
\title{Enhancing Semantic Clustering for Uncertainty Quantification \& Conformal Prediction by LLMs}
\title{Acknowledging Uncertainty in LLMs for Reliable Generative AI}
\title{Addressing Uncertainty in LLMs to Enhance Reliability in Generative AI}

\author{
 \textbf{Ramneet Kaur\textsuperscript{1}},
 \textbf{Colin Samplawski\textsuperscript{1}},
 \textbf{Adam D. Cobb\textsuperscript{1}},
 \textbf{Anirban Roy\textsuperscript{1}},
 \textbf{Brian Matejek\textsuperscript{1}},\\
 \textbf{Manoj Acharya\textsuperscript{1}},
 \textbf{Daniel Elenius\textsuperscript{1}},
 \textbf{Alexander M. Berenbeim\textsuperscript{2}},
 \textbf{John A. Pavlik\textsuperscript{2}},\\
 \textbf{Nathaniel D. Bastian\textsuperscript{2}},
 \textbf{Susmit Jha\textsuperscript{1}}
\\
 \textsuperscript{1}Neuro-symbolic Computing and Intelligence, SRI, Menlo Park, USA \\
 \textsuperscript{2}Army Cyber Institute, United States Military Academy, West Point, NY USA
\\
 \small{
   \textbf{Correspondence:} \href{mailto:email@domain}{ramneet.kaur@sri.com}
 }
}

\newcommand{\newMetricName}{AURAC} 

\begin{document}

\maketitle

\begin{abstract}

  In this paper, we present a dynamic semantic clustering approach inspired by the Chinese Restaurant Process, aimed at addressing uncertainty in the inference of Large Language Models (LLMs). We quantify uncertainty of an LLM on a given query by calculating entropy of the generated semantic clusters. Further, we propose leveraging the (negative) likelihood of these clusters as the (non)conformity score within Conformal Prediction framework, allowing the model to predict a set of responses instead of a single output, thereby  accounting for uncertainty in its predictions. We demonstrate the effectiveness of our uncertainty quantification (UQ) technique on two well-known question-answering benchmarks, COQA and TriviaQA, utilizing two LLMs—Llama-2-13b and Mistral-7b. Our approach achieves state-of-the-art (SOTA) performance in UQ, as assessed by metrics such as AUROC, AUARC, and \newMetricName. The proposed conformal predictor is also shown to produce smaller prediction sets while maintaining the same probabilistic guarantee of including the correct response, in comparison to existing SOTA conformal prediction baseline. Our code is publicly accessible at \url{https://shorturl.at/7yHSq}.
\end{abstract}


\section{Introduction}
\label{sec:intro}
%


Large language models (LLMs) are increasingly being utilized for various tasks, including open-world question answering. However, these models are known to exhibit hallucinations, confidently producing incorrect information or relying on faulty reasoning. Quantifying the uncertainty associated with an LLM for a given input offers a practical method to account for model's reliability on its response. When there is a strong correlation between LLM's accuracy and the computed uncertainty, one can effectively use uncertainty quantification (UQ) approach to determine when to place their trust in the model's responses. 


The challenge of UQ for LLMs in generative context differs significantly from that encountered in regression or classification tasks. While the latter has been well-studied \cite{guo2017calibration,jha2019attribution,xiao2019quantifying,hu2021uncertainty,Magesh2023PrincipledOD}, UQ for generative models presents unique challenges due to the free-form nature of the responses of varying lengths. Additionally, the syntactic similarity of generated sequences may not necessarily align with their semantic similarity. 

Self-consistency theory can be employed to measure LLM's uncertainty on an input query by comparing semantic context of multiple outputs sampled from the LLM for that query~\citep{wang2022self}.\footnote{Higher semantic consistency among multiple outputs would imply less uncertainty.} 
Building on recent observations by~\citet{lin, kuhn}, we use embedding-based semantic similarity to group sampled outputs into semantic clusters, and then quantify LLM's uncertainty by entropy of these clusters. Specifically, we propose a new dynamic semantic clustering algorithm based on the sequential distance dependent Chinese Restaurant Process (DDCRP)~\cite{blei2011distance, tuncer2016sequential} for quantifying LLM's uncertainty via entropy of the generated clusters.


Furthermore, we present a novel approach towards trustworthy inference from LLMs that combines the model's uncertainty with conformal prediction. Specifically, we propose to use (negative) likelihood of the clusters generated via DDCRP, as the (non)conformity score in the conformal prediction framework~\citep{cp}. This framework takes into account prediction uncertainty by generating a set instead of a single prediction, where the output set is guaranteed to contain the correct label (or answer in our case) with a certain confidence level. LLMs can generate free-form responses for question answering and so, the conformal prediction set in our approach is different from the usual setting of being a subset of known finite set of labels. We use semantic similarity to identify different or semantically diverse responses while constructing the conformal set. This use of conformal prediction by LLMs in question answering is the key novelty of our approach that aims to combine LLMs uncertainty on an input query with conformal prediction.

We demonstrate the efficacy of our UQ approach on two question-answering benchmarks, COQA and TriviaQA, using two LLMs, Llama-2-13b and Mistral-7b, achieving state-of-the-art (SOTA) results in most test cases using AUROC, AUARC and \newMetricName\ as metrics. We also show that our conformal predictor generates smaller prediction sets for the same probabilistic guarantees of including correct response compared to the SOTA conformal prediction baseline by~\citet{conformal_lang_modeling}. This highlights not only the validity but also usefulness of our method in practical applications.

\section{Related Work}
\label{sec:rel_work}

\textbf{Uncertainty quantification (UQ)} for LLMs 
has received significant attention over the last few years. 
One approach is to explicitly query the model for the correctness 
probability~\citep{self_prob}. Another approach relies on utilizing the log-likelihood~\citep{log_likelihood} associated with its generated response by taking a product or an average or other statistical aggregation over the generated tokens. LLMs are known to be not well-calibrated~\citep{reducing_overconf_cali} and consequently, methods for calibrating LLMs~\citep{calibrating_long} have also been proposed.
Semantic entropy or predictive uncertainty that measures the (in)consistency among multiple responses has been proposed as a metric for UQ of LLMs~\citep{kuhn, lin}. We also use semantic entropy for UQ but adopt a novel semantic clustering approach and empirically demonstrate its effectiveness.

\noindent \textbf{Conformal prediction (CP)}~\citep{cp} has been used 
for deploying deep learning models~\citep{idecode, haroush2021statistical, codit, kaur2024out, yang2024memory} in high-assurance applications wherein the model predicts a set instead of a single prediction such that one of the responses in the set is guaranteed to be correct with a probability higher than the user specified significance level. 
In the context of LLMs, conformal prediction has been used for providing coverage guarantees~\citep{UQ_benchmark, conformal_lang_modeling}. 
~\citet{UQ_benchmark} concentrate on classification settings and propose non-conformity scores in the CP framework accordingly.
In contrast, we focus on generative setting for LLMs in applications such as question-answering.~\citet{conformal_lang_modeling} propose generating diverse prediction sets based on the quality of individual responses, and a set scoring function. They utilize CP to derive hyperparameters ($\lambda$s) for diversity, quality, and set scoring function in their algorithm for coverage guarantees. We, instead propose, using (negative) likelihood of clusters representing semantically diverse responses, as the (non)conformity score in the CP framework for generating sets with coverage guarantees, and compare our results with~\citet{conformal_lang_modeling}'s approach.

\section{Clustering by Semantic Equivalence}
\label{sec:tech}
In this section, we introduce the proposed clustering approach, and then describe its usage in uncertainty quantification and building the conformal predictor for LLMs.

\textit{Semantic entropy} over $|C|$ equivalence classes (or clusters) of multiple responses sampled from an LLM on an input query $x$ has been used as the \textit{UQ measure} for the LLM on $x$~\citep{kuhn}:
\begin{equation*}
    SE(x) = -\sum_c p(c|x)\log p(c|x),
\end{equation*}
where the probability of an equivalence class (corresponding to a cluster of embeddings), $c$, conditioned on $x$, is given by $p(c|x) = \sum_{\mathbf{s}\in c} p(\mathbf{s}|x)$ and $\mathbf{s}\in c$ denotes a sentence (or response) in an equivalence class $c \in C$. The probability of each sentence is given by the standard product of the conditional token probabilities:
\begin{equation*}
    p(\mathbf{s}|x) = \prod_i p(s_i|s_{<i},x).
\end{equation*}
The result of using semantic entropy is to sharpen $p(c|x)$ when there are many responses with the same meaning, and therefore reduce the predicted entropy. 

To generate clusters containing semantically equivalent responses, we need to define a function that checks semantic similarity between responses.~\citet{kuhn} choose the conservative approach of using Deberta (Natural Language Inference) model \cite{he2020deberta} to only define two sentences to be semantically equivalent if and only if entailment was classified in both directions. 
Entailment in both directions is not necessary when one response includes additional information than another but they both contain the same relevant semantic information. For instance, let us consider the following input query: ``What is the capital of France?'' with the first response $\mathbf{s_1}=$ ``Paris'', and the second response $\mathbf{s_2}=$ ``Paris, the capital city of France, is one of the most iconic and romantic cities in the world. It has a rich history dating back more than 2,000 years''. Although, both the responses are semantically equivalent with respect to the input query, $\mathbf{s_2}$ entails $\mathbf{s_1}$ but vice versa is not true. Enforcing entailment in both directions can, therefore, lead to formation of more clusters than required with semantically equivalent information distributed in different clusters.


Thus, our approach computes the probability that two responses, $\mathbf{s}_i$ and $\mathbf{s}_j$, are in the same equivalence class (that is, the same semantic cluster), by taking the maximum entailment score output by Deberta, $p_{ij} = \max (p(\mathbf{s}_i\vdash \mathbf{s}_j), (p(\mathbf{s}_j\vdash \mathbf{s}_i)) $.
This choice of taking a maximum can be viewed as a quantitative disjunction of entailment in either direction. 
We then use this approach to build the score that a response, $\mathbf{s}_j$, belongs to an equivalence class, $c$, by using the average probability across all the cluster members:
\begin{equation}\label{eq:p_cluster}
    w(\mathbf{s}_j \in c) = \frac{1}{|c|}\sum_{\mathbf{s}_i \in c} p_{ij}
\end{equation}
This is also novel from the existing approaches for semantic clustering where $\mathbf{s_j}$ is assigned to the cluster $c$ if it is entailed in both directions by only one member of $c$. As observed in our qualitative evaluation (and reported in Appendix~\ref{app:qualcluster}), we postulate that this could be the reason for assigning semantically irrelevant responses to the same cluster because it is easier to incorrectly assign responses to the same cluster if we rely on only one member instead of all members in the cluster. In contrast, we take an average over all existing members of a cluster making our assignment more robust. 

A na\"ive approach would be to greedily assign $\mathbf{s_j}$ to a cluster with the highest probability. However, this is not sufficient since we need a mechanism for forming new equivalence classes (or clusters). Given a set of responses, we need to decide when to form a new cluster and when 
to assign $\mathbf{s_j}$ to
an existing cluster $c$ that has the highest score 
$ w(\mathbf{s_j} \in c)$.
We use the same mechanism as the Distance Dependent Chinese Restaurant Process (DDCRP) for iterative clustering~\citep{blei2011distance, tuncer2016sequential}. The motivation for using DDCRP is due to its probabilistic and dynamic nature for deciding on the formation of a new cluster or membership in an existing cluster based on semantic equivalence. 
Following DDCRP, the score for a sentence $\mathbf{s_j}$ in a new cluster $c^*$ is defined as:
\begin{equation}
    w(\mathbf{s_j} \in c^*) = \frac{\alpha}{\alpha + |C|},
    \label{prob_new_cluster}
\end{equation}
where $|C|$ is the current number of clusters and $\alpha > 0$ is the rate parameter, i.e. prior over forming new clusters. The final (softmax) probabilities of assigning $\mathbf{s_j}$ to an existing cluster ($c_i$) and new cluster ($c^*$) are respectively given by:
\begin{align*}
  \text{Softmax}(z_i) = \frac{e^{z_i}}{\sum_j e^{z_j} + e^{z^*}} \notag \\
  \text{Softmax}(z^*) = \frac{e^{z^*}}{\sum_j e^{z_j} + e^{z^*}} \notag,
  \label{norm_probs}
  \numberthis
\end{align*}

where $z_i = w(s \in c_i), \text{and } z^* = w(s \in c^*).$
Since the equivalence class assignment of the new response is related to semantic equivalence with the existing responses, we can map our clustering algorithm as an instance of sequential DDCRP.

Alg.~\ref{alg:clustering} is the proposed clustering approach for iterative clustering of semantically equivalent responses. It is executed in a sequential fashion, starting with an empty set of clusters. After the first response from the LLM, we compute the score for forming a new cluster as: $\frac{\alpha}{\alpha + 0}=1$. This results in a new cluster probability of 1, leading a new cluster to be formed deterministically. In subsequent rounds, we compute the per cluster assignment scores for the new response as per Equations~\eqref{eq:p_cluster} and~\eqref{prob_new_cluster} for existing and new cluster respectively. We, then, compute the softmax probabilities for these scores from Equation~\eqref{norm_probs}, and assign the new response to a cluster (either existing cluster $c_i$ or a new cluster $c^*$) with the maximum probability. 


\begin{algorithm}
\caption{Clustering by Semantic Equivalence}
\label{alg:clustering}
\begin{algorithmic}[1]
\STATE{\bfseries Input:} query $x$, LLM model $M$
\STATE{\bfseries Parameter:} rate parameter $\alpha > 0$, number of responses $N$
\STATE{\bfseries Output:} clusters $C$
\STATE{\bfseries Initialize:}  $C \leftarrow \emptyset$ 
\FOR{$i = 1$ to $N$}
\STATE $\mathbf{s}_i = M(x)$ \COMMENT{generate with LLM}
\STATE $scores \leftarrow \mathbf{0}_{|C| + 1}$
\FOR{$c_j$ in $C$}
\STATE $scores[j] \leftarrow w(\mathbf{s}_i \in c_j)$
\ENDFOR
\STATE $scores[-1] \leftarrow \frac{\alpha}{\alpha + |C|}$
\STATE $probs \leftarrow \text{Softmax}(scores)$
\STATE $k \leftarrow argmax(probs)$ \COMMENT{cluster assignment}
\IF{$k == |C|+1$}
\STATE $C \leftarrow C \cup \{\mathbf{s}_i\}$ \COMMENT{new cluster}
\ELSE
\STATE $c_k \leftarrow c_k \cup \{\mathbf{s}_i\}$
\ENDIF
\ENDFOR
\STATE \RETURN $C$
\end{algorithmic}
\end{algorithm}

In all our experiments, we use the rate parameter of $\alpha=0.5$. We performed a grid search over  $\alpha \in [0.2,0.3,\ldots,0.7]$ and observed very little variance in performance. We set the value of number of responses $N$ is set to $20$, which is consistent with the existing work~\citep{kuhn, lin, conformal_lang_modeling}.

\subsection{Uncertainty Quantification} 
Different clusters containing semantically diverse responses to an input query can be used to quantify uncertainty of the LLM on the query. Similar to~\citet{kuhn}, we also use entropy of the generated clusters from Alg.~\ref{alg:clustering} as a measure of UQ for LLMs on a given query. Both qualitative and quantitative results indicate that the proposed clustering approach yields better results than Kuhn's baseline.

\subsection{Conformal Set Prediction}
\label{sec:cp}
For each cluster $c$, we have $p(c|x)$. We, therefore, use the negative log probability, $\log [1/p(c|x)]$, of the individual clusters, as the non-conformity score  
in conformal prediction (CP) framework~\citep{cp}, for generating prediction sets. 
Non-conformity scores of calibration datapoints is used to build a reference empirical distribution to compare against when building the prediction set. Specifically, depending on the desired significance level, $\epsilon$, prediction set is generated by comparing scores for the test clusters with a threshold from the empirical distribution: non-conformity score of calibration set at $(1-\epsilon)^{th}$ quantile of the distribution.\footnote{Calibration set consists of only those clusters whose all responses are accurate.} Intuitively, clusters with low negative log probability (or high likelihood) are more likely to be included in prediction sets compared to clusters with high negative log probability (or low likelihood). If an LLM outputs many semantically equivalent responses, then we expect the cluster's $\log [1/\sum_{\mathbf{s}\in c} p(\mathbf{s}|x)]$ to decrease due to the summation over the sentence probabilities by sharpening the cluster probability.

The use of CP for constructing the prediction sets gives us coverage guarantees on the true answer in the set with the probability greater than or equal to $1-\epsilon$~\citep{icp}. Alg.~\ref{alg:cp_sets} is the proposed CP algorithm for generating prediction sets with coverage guarantees.

For an input query $x$ (from test or calibration set), clusters are generated via Alg.~\ref{alg:clustering}. We use negative log probability (nlp = $\log [1/p(c|x)]$) of each generated cluster ($c$) for $x$ as the non-conformity score for the cluster. For the desired significance level $\epsilon \in (0, 1)$, the prediction threshold $\tau$ is decided as the score at $(1-\epsilon)^{th}$ quantile of the empirical distribution of non-conformity scores for the calibration clusters. Assuming all responses are semantically equivalent in a cluster, a single response from the test cluster ($c$) is added to the prediction set if its non-conformity score (nlp($c$)) is below the prediction threshold $\tau$. In our experiments, prediction set is constructed from the first response in the qualified test cluster.

\begin{algorithm}
    \caption{Conformal Prediction Sets by LLM}
    \label{alg:cp_sets}
    \begin{algorithmic}[1]
        \STATE{\bfseries Input:} 
        query $x$, LLM model $M$,
        prediction threshold $\tau$ from $(1-\epsilon)^{th}$ quantile of the empirical distribution of calibration set non-conformity scores
        \STATE {\bfseries Output:} Prediction Set $\mathcal{O}$ with predictions on $x$ by $M$ s.t. $ Pr. (\text{correct answer} \in \mathcal{O}) \geq 1-\epsilon$
        \STATE $C =$ set of clusters from Alg.~\ref{alg:clustering} on ($x$, $M$)
        \STATE $S =$ set of non-conformity scores for the generated clusters: $\{\forall c \in C: \text{nlp}(c)\}$
        \STATE $\mathcal{O} = \{\text{a response from } C[i] \text{ s.t. } S[i] \leq \tau: i = 1,\ldots,|C| \}$
        \STATE return $\mathcal{O}$
    \end{algorithmic}
\end{algorithm}


\section{Experimental Results}
\label{sec:exp}
The experimental evaluation focuses on two research questions. \textbf{RQ1:} Does the novel semantic clustering approach inspired from CRP improve the UQ of LLMs? \textbf{RQ2:} How does Alg~\ref{alg:cp_sets} perform compared to the CP baseline on LLMs for free form generative responses? 


\noindent {\textbf{Datasets and Models}}. We use two question-answer datasets: COQA~\citep{coqa} and TriviaQA~\citep{triviaqa}, over which we compare the performance of two LLMs, Llama-2-13b: non-instruct model~\citep{llama-2}, and Mistral-7b: instruct model~\citep{mistral}. 
Following existing literature~\citep{lin, kuhn}, 
we deploy three evaluation methods: (1) We query GPT-4~\citep{achiam2023gpt} by asking it to provide a rating on whether a response is correct with a value between 0 and 1, and label the response as correct if its rating $>0.7$; (2) RougeL score~\citep{rouge} with a threshold  $>0.3$; (3) Deberta~\citep{he2020deberta} to check for entailment of correct answer in the generated response.



\subsection{UQ Performance}
\label{sec:uq_perf}
We report Area Under Accuracy-Rejection Curve (AUARC)~\citep{auarc}, and Area Under Rejection-Accuracy Curve (\newMetricName) for comparing our performance on UQ with \citet{kuhn}'s, and \citet{lin}'s approaches. For~\citet{lin}'s approach, we use EigV as their UQ metric\footnote{They also propose `Ecc', and `Deg' as other UQ metrics. Consistent with their paper, we found that the best results in most of the cases are with `EigV' metric, and therefore we compare our results with this UQ metric.}. 
While AUARC has been used as an evaluation metric previously~\citep{lin}, we include \newMetricName\ as a new metric. 
AUARC is an indicator of the accuracy of accepted (or highly certain) samples, and \newMetricName\ is an indicator of the accuracy of rejected (or highly uncertain) samples. 
In addition to AUARC, \newMetricName\ also indicates calibration of the UQ metric: we would like the accuracy of the model on the rejected samples by the UQ metric to be as low as possible, i.e. not rejecting samples on which LLMs are accurate.

The results are reported in Tables~\ref{tab:all_auarc_results}, and~\ref{tab:all_rej_acc_results}. Similar to~\citet{kuhn}'s, we also report our results with the UQ score unnormalized (Unnorm)/normalized (Norm) on the response's length. We outperform the baseline by~\citet{kuhn} in all the test cases, indicating that the proposed clustering approach performs better in UQ. 
We achieve competitive results in comparison to the current SOTA by~\citet{lin} by outperforming them in most cases. We also report AUROC, and compare with other baselines in Appendix~\ref{app_sec_all_uq_results}.
\begin{table*}[!t]
    \centering
    \setlength{\tabcolsep}{2pt}
    \resizebox{1\columnwidth}{!}{
    \begin{tabular}{c|c||c|c|c|c||c|c|c|c}
    \hline
    \multicolumn{2}{c||}{} & \multicolumn{4}{c||}{\textbf{COQA} Dataset} & \multicolumn{4}{c}{\textbf{TriviaQA} Dataset} \\
    \hline
        \textbf{Model} & \textbf{Eval.} & \textbf{Model} & \textbf{Sem. Ent.} & \textbf{EigV} & \textbf{Ours} & \textbf{Model} & \textbf{Sem. Ent.} & \textbf{EigV} & \textbf{Ours} \\
        & & \textbf{Acc.} & Unnorm/Norm & & Unnorm/Norm & \textbf{Acc.} & Unnorm/Norm & & Unnorm/Norm
        \\
        \hline
        Llama-13b & GPT-4 & 73.22 & 85.81/86.44 & \textbf{88.03} & 86.35/\underline{87.47} & 67.03 & 88.13/87.94 & \textbf{88.84} & 88.33/\underline{88.54} \\
        Mistral-7b & GPT-4 & 73.38 &  81.91/82.68 & \underline{82.82} & 82.22/\textbf{82.95} & 60.68 & 80.99/\underline{81.40} & \textbf{82.03} & 81.23/\textbf{82.03} \\
        \hline
        Mean & GPT-4 & 73.30 & 83.86/84.56 & \textbf{85.43} & 84.29/\underline{85.21}& 63.86 & 84.56/84.67 & \textbf{85.44} & 84.78/\underline{85.29} \\
        \hline
        Llama-13b & RougeL & 72.75 &  86.03/87.05 & \underline{87.92} & 86.84/\textbf{88.34}& 64.60 & 85.62/85.19 & 85.76 & \underline{85.86}/\textbf{85.87} \\
        Mistral-7b & RougeL & 44.74 & \underline{64.37}/62.93  & 63.43 & \textbf{64.60}/63.48 & 42.33 & \underline{70.18}/68.13 & 69.41 & \textbf{70.26}/68.81 \\
        \hline
        Mean & RougeL & 58.75 & 75.20/74.99 & 75.65  & \underline{75.72}/\textbf{75.91}  & 53.47 & \underline{77.90}/76.66 & 77.59 & \textbf{78.06}/77.34 \\
        \hline
        Llama-13b & Deberta & 63.74 & 80.21/79.48 & \textbf{82.68} & 81.04/\underline{81.37} & 63.33 & 84.92/84.34  & \textbf{85.60} & \underline{85.23}/85.13 \\
        Mistral-7b & Deberta & 11.23 & \underline{23.56}/20.71 & 20.88 & \textbf{23.53}/21.05 & 33.92 & \underline{62.29}/59.53 & 60.39 & \textbf{62.37}/60.16 \\
        \hline
        Mean & Deberta & 37.49 & \underline{51.89}/50.10 & 51.78 & \textbf{52.29}/51.21 & 48.63 & \underline{73.61}/71.94 & 73.00 & \textbf{73.80}/72.65 \\
        \hline
    \end{tabular}
    }
    \caption{AUARC ($\uparrow$) results in comparison to~\citet{kuhn}'s Semantic Entropy (Sem. Ent.) UQ metric, and SOTA EigV metric by~\citet{lin}. Best results are in bold and second best are underlined.}
    \label{tab:all_auarc_results}
\end{table*}

\begin{table*}[!t]
    \centering
    \setlength{\tabcolsep}{2pt}
    \resizebox{1\columnwidth}{!}{
    \begin{tabular}{c|c||c|c|c|c||c|c|c|c}
    \hline
    \multicolumn{2}{c||}{} & \multicolumn{4}{c||}{\textbf{COQA} Dataset} & \multicolumn{4}{c}{\textbf{TriviaQA} Dataset} \\
    \hline
        \textbf{Model} & \textbf{Eval.} & \textbf{Model} & \textbf{Sem. Ent.} & \textbf{EigV} & \textbf{Ours} & \textbf{Model} & \textbf{Sem. Ent.} & \textbf{EigV} & \textbf{Ours} \\
        & & \textbf{Acc.} & Unnorm/Norm & & Unnorm/Norm & \textbf{Acc.} & Unnorm/Norm & & Unnorm/Norm
        \\
        \hline
        Llama-13b & GPT-4 & 73.22  & 58.97/56.90 & \textbf{54.63} &  58.42/\underline{55.32} & 67.03 & 40.09/40.27 & \underline{39.42} & 39.92/\textbf{39.38} \\
        Mistral-7b & GPT-4 & 73.38 & 63.06/62.02 & \textbf{59.83} & 62.77/\underline{61.41} & 60.68 & 35.57/35.04 & \textbf{33.19} & 35.13/\underline{33.29}\\
        \hline
        Mean & GPT-4 & 73.30  & 61.02/59.46 & \textbf{57.23} & 60.60/\underline{58.37} & 63.86& 37.83/37.66 &  \textbf{36.31} & 37.53/\underline{36.34} \\
        \hline
        Llama-13b & RougeL & 72.75 & 56.75/55.53 & \underline{53.78} & 55.78/\textbf{52.65}& 64.60 & 39.12/39.39 & 38.93 &  \underline{38.81}/\textbf{38.35} \\
        Mistral-7b & RougeL & 44.74 & 27.62/29.65 & \textbf{27.12} & \underline{27.37}/28.26 & 42.33& 17.15/19.56 & \underline{17.06} & \textbf{16.95}/18.11 \\
        \hline
        Mean & RougeL & 58.75 & 42.19/42.59 & \textbf{40.45} & 41.58/\underline{40.46} & 53.47& 28.14/29.48 & \underline{28.00} & \textbf{27.88}/28.23 \\
        \hline
        Llama-13b & Deberta &  63.74 & 46.07/46.91 & \textbf{42.04} & 45.07/\underline{43.56} & 63.33 & 37.23/37.94 & \textbf{36.70} & \underline{36.88}/36.84  \\\
        Mistral-7b & Deberta & 11.23 & \underline{3.84}/5.70 & 4.13 & \textbf{3.82}/5.00 & 33.92 & \underline{11.00}/13.54 &   11.35 & \textbf{10.89}/12.45\\
        \hline
        Mean & Deberta & 37.49 & 24.96/26.31 & \textbf{23.09} & \underline{24.45}/24.28 & 48.63 & 24.12/25.74  & \underline{24.03} & \textbf{23.89}/24.65\\
        \hline
    \end{tabular}
    }
    \caption{\newMetricName\ ($\downarrow$) results in comparison to~\citet{kuhn}'s Semantic Entropy (Sem. Ent.) UQ metric, and SOTA EigV metric by~\citet{lin}. Best results are in bold and second best are underlined.}
    \label{tab:all_rej_acc_results}
\end{table*}


 \begin{figure*}[!t]
        \includegraphics[width=1\columnwidth]{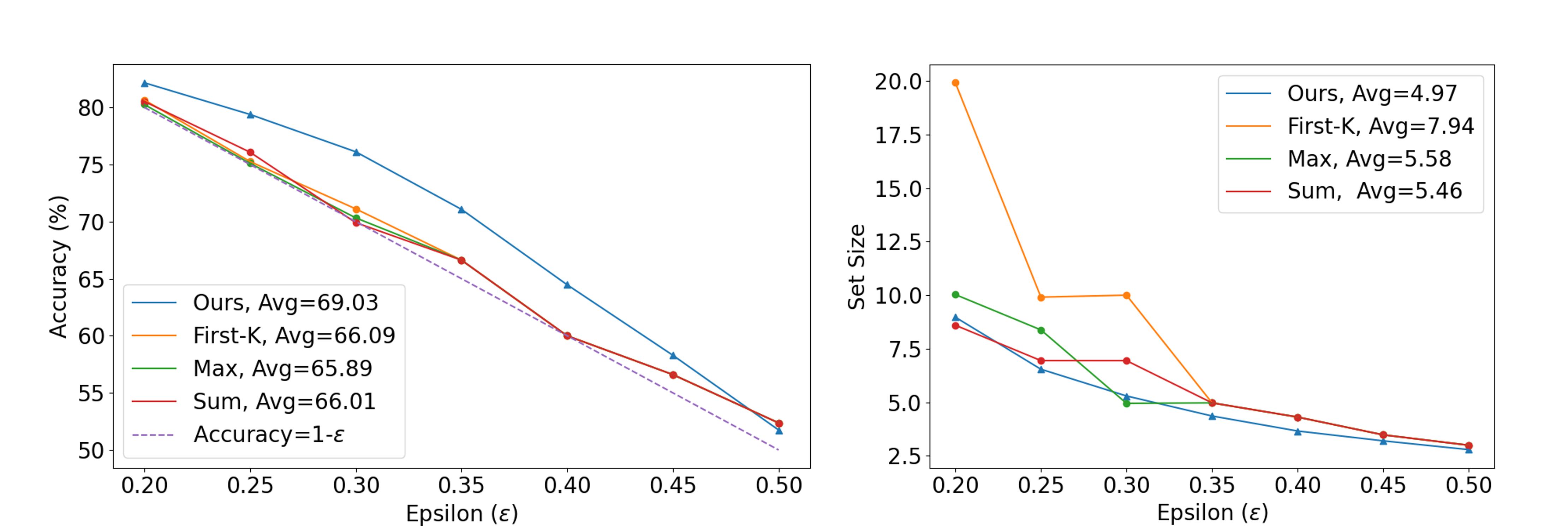}
    \caption{Comparison of Accuracy (left) and Set Size of prediction sets (right) with Conformal Prediction baseline~\citep{conformal_lang_modeling}.}
\label{fig:cp_comp_curves}
\end{figure*}

\subsection{Conformal Prediction Results}
The desired properties of a prediction set is that the accuracy of the set should be as high as possible with a smaller set size. So, here we report accuracy and set size as the evaluation metrics.
\subsubsection{Comparison with the CP Baseline}
Fig.~\ref{fig:cp_comp_curves} shows Alg.~\ref{alg:cp_sets} results in comparison with the existing baseline by~\citet{conformal_lang_modeling} on using CP for generating prediction sets with coverage guarantees. 

The coverage guarantee (or guarantee of the correct answer contained in the prediction set by CP) is expected to be $\geq (1-\epsilon)$. So, as the value of $\epsilon$ increases, the accuracy and the set size is expected to decrease. This is what we observe for both approaches: ours and the baseline, with both the approaches satisfying the coverage guarantees.~\citet{conformal_lang_modeling} report results with different variations of their proposed algorithm (Algorithm 1 of their paper) in terms of the set scoring function ($\mathcal{F}$): First-$K$, Max, and Sum, and on \textbf{TriviaQA with Llama-2-13b}. We outperform these results for all the three variations on both evaluation metrics. 

\subsubsection{Experiments on COQA}
\begin{figure*}[!t]
        \centering
        \includegraphics[width=1\columnwidth]{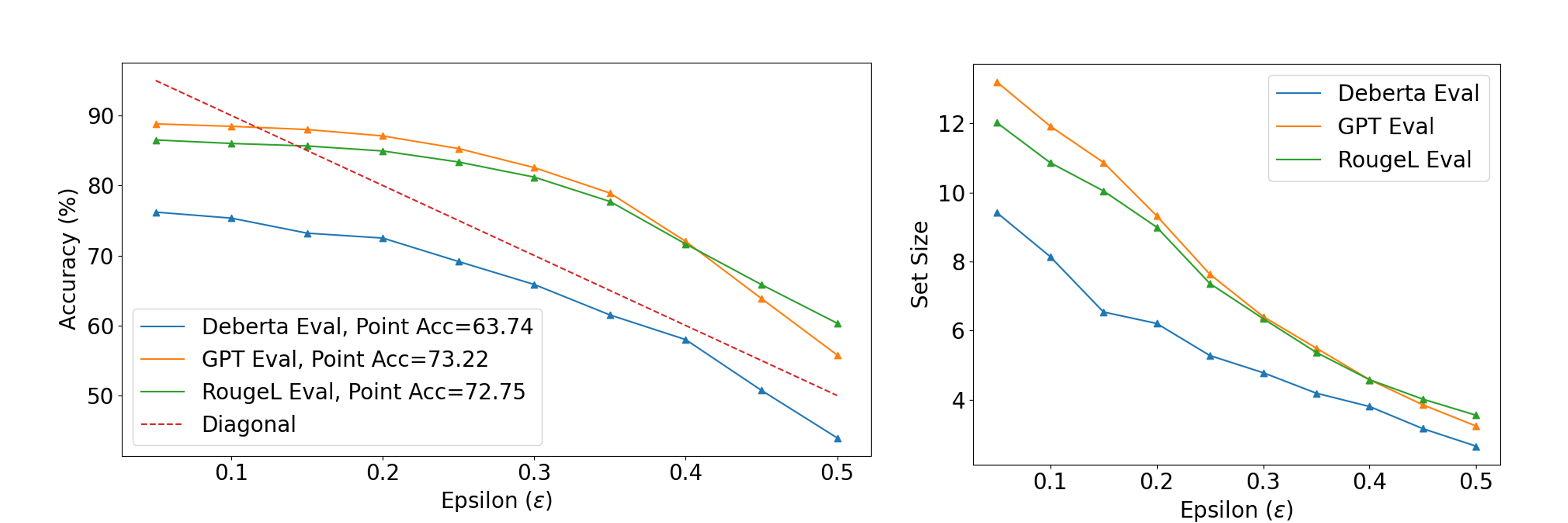}
    \caption{Alg.~\ref{alg:cp_sets}'s Accuracy (left) and Set Size (right) evaluation on COQA for Llama-13b.}
\label{fig:cp_acc_ss_curves_coqa}
\end{figure*}

We also evaluate our Alg.~\ref{alg:cp_sets}'s performance on COQA. Figure~\ref{fig:cp_acc_ss_curves_coqa} shows these results for both accuracy and set size and with all the three GT evaluation approaches on COQA: GPT-4, RougeL, and Deberta. 

Here, we also report the point accuracy, which is the average accuracy of the individual $N=20$ generations. For $\epsilon \leq 0.35$, the prediction set accuracy is always higher than the point accuracy. Consistent with the results on TriviQA, here also the value of accuracy and set size decreases with the increase in the value of $\epsilon$. GPT and RougeL evaluations satisfy the coverage guarantee $\forall \epsilon \geq 0.15$. Even though conformal prediction provides a rigorous theoretical guarantee, deviations from the coverage guarantee can occur in practice due to limited sample variability in the caliibration set~\citep{angelopoulos2021gentle}. This justifies the accuracy results with Deberta Evaluation and the other two evaluations with $\epsilon < 0.15$.

One difference to note here from the TriviaQA experiments is in the set of $\epsilon$ values. For TriviaQA, we reported results with $\epsilon \in \{0.2,\ldots,0.5\}$, and for COQA we reported results with $\epsilon \in \{0.1,0.2,\ldots,0.5\}$. This is because we were getting `nan' at $\epsilon=0.1$ from the Quach's baseline on TriviaQA. While further investigation on their code, we figured it out that they are hard-coding the value as `nan' where the search for their algorithm's hyperparameters ($\lambda$s) might be failing.

\section{Conclusion}
This paper takes a step towards enhancing reliability in generative AI by addressing uncertainty in LLMs on a given query. Our approach focuses on the semantic equivalence of responses generated by an LLM when prompted multiple times for the input query. It is based on the idea that an LLM is expected to be accurate if it consistently generates semantically similar outputs when prompted multiple times with the same input. The underlying assumption here is that consistency can serve as an indicator of accuracy. Testing this hypothesis is one of the future works that we intend to investigate. 
Additionally, we aim to explore alternative scoring methods beyond the probability of response—calculated as the product of conditional token probabilities— used to determine the likelihood of semantic clusters. This is important because response probability can be sensitive to its length, which poses a significant challenge when dealing with free-form generations produced by LLMs.

\section*{Acknowledgments}
This work was supported in part by the United States Air Force and Defense Advanced Research Projects Agency (DARPA) under Contract No.FA8750-23-C-0519, the Defense Advanced Research Projects Agency (DARPA) under Agreement No. HR0011-24-9-0424 and Defense Logistics Agency (DLA) and the Advanced Research Projects Agency for Health (ARPA-H) under Contract Number SP4701-23-C-0073. Any opinions, findings and conclusions or recommendations expressed in this material are those of the authors and do not necessarily reflect the Department of Defense, DARPA, DLA, ARPA-H or the United States Government.

\newpage
\bibliography{main}


\begin{thebibliography}{32}


\ifx \showCODEN    \undefined \def \showCODEN     #1{\unskip}     \fi
\ifx \showDOI      \undefined \def \showDOI       #1{#1}\fi
\ifx \showISBNx    \undefined \def \showISBNx     #1{\unskip}     \fi
\ifx \showISBNxiii \undefined \def \showISBNxiii  #1{\unskip}     \fi
\ifx \showISSN     \undefined \def \showISSN      #1{\unskip}     \fi
\ifx \showLCCN     \undefined \def \showLCCN      #1{\unskip}     \fi
\ifx \shownote     \undefined \def \shownote      #1{#1}          \fi
\ifx \showarticletitle \undefined \def \showarticletitle #1{#1}   \fi
\ifx \showURL      \undefined \def \showURL       {\relax}        \fi
\providecommand\bibfield[2]{#2}
\providecommand\bibinfo[2]{#2}
\providecommand\natexlab[1]{#1}
\providecommand\showeprint[2][]{arXiv:#2}

\bibitem[Achiam et~al\mbox{.}(2023)]%
        {achiam2023gpt}
\bibfield{author}{\bibinfo{person}{Josh Achiam}, \bibinfo{person}{Steven Adler}, \bibinfo{person}{Sandhini Agarwal}, \bibinfo{person}{Lama Ahmad}, \bibinfo{person}{Ilge Akkaya}, \bibinfo{person}{Florencia~Leoni Aleman}, \bibinfo{person}{Diogo Almeida}, \bibinfo{person}{Janko Altenschmidt}, \bibinfo{person}{Sam Altman}, \bibinfo{person}{Shyamal Anadkat}, {et~al\mbox{.}}} \bibinfo{year}{2023}\natexlab{}.
\newblock \showarticletitle{Gpt-4 technical report}.
\newblock \bibinfo{journal}{\emph{arXiv preprint arXiv:2303.08774}} (\bibinfo{year}{2023}).
\newblock


\bibitem[Angelopoulos and Bates(2021)]%
        {angelopoulos2021gentle}
\bibfield{author}{\bibinfo{person}{Anastasios~N Angelopoulos} {and} \bibinfo{person}{Stephen Bates}.} \bibinfo{year}{2021}\natexlab{}.
\newblock \showarticletitle{A gentle introduction to conformal prediction and distribution-free uncertainty quantification}.
\newblock \bibinfo{journal}{\emph{arXiv preprint arXiv:2107.07511}} (\bibinfo{year}{2021}).
\newblock


\bibitem[Balasubramanian et~al\mbox{.}(2014)]%
        {cp}
\bibfield{author}{\bibinfo{person}{Vineeth Balasubramanian}, \bibinfo{person}{Shen-Shyang Ho}, {and} \bibinfo{person}{Vladimir Vovk}.} \bibinfo{year}{2014}\natexlab{}.
\newblock \bibinfo{booktitle}{\emph{Conformal prediction for reliable machine learning: theory, adaptations and applications}}.
\newblock \bibinfo{publisher}{Newnes}.
\newblock


\bibitem[Blei and Frazier(2011)]%
        {blei2011distance}
\bibfield{author}{\bibinfo{person}{David~M Blei} {and} \bibinfo{person}{Peter~I Frazier}.} \bibinfo{year}{2011}\natexlab{}.
\newblock \showarticletitle{Distance dependent Chinese restaurant processes.}
\newblock \bibinfo{journal}{\emph{Journal of Machine Learning Research}} \bibinfo{volume}{12}, \bibinfo{number}{8} (\bibinfo{year}{2011}).
\newblock


\bibitem[Guo et~al\mbox{.}(2017)]%
        {guo2017calibration}
\bibfield{author}{\bibinfo{person}{Chuan Guo}, \bibinfo{person}{Geoff Pleiss}, \bibinfo{person}{Yu Sun}, {and} \bibinfo{person}{Kilian~Q Weinberger}.} \bibinfo{year}{2017}\natexlab{}.
\newblock \showarticletitle{On calibration of modern neural networks}. In \bibinfo{booktitle}{\emph{International conference on machine learning}}. PMLR, \bibinfo{pages}{1321--1330}.
\newblock


\bibitem[Haroush et~al\mbox{.}(2021)]%
        {haroush2021statistical}
\bibfield{author}{\bibinfo{person}{Matan Haroush}, \bibinfo{person}{Tzviel Frostig}, \bibinfo{person}{Ruth Heller}, {and} \bibinfo{person}{Daniel Soudry}.} \bibinfo{year}{2021}\natexlab{}.
\newblock \showarticletitle{A statistical framework for efficient out of distribution detection in deep neural networks}.
\newblock \bibinfo{journal}{\emph{arXiv preprint arXiv:2102.12967}} (\bibinfo{year}{2021}).
\newblock


\bibitem[He et~al\mbox{.}(2020)]%
        {he2020deberta}
\bibfield{author}{\bibinfo{person}{Pengcheng He}, \bibinfo{person}{Xiaodong Liu}, \bibinfo{person}{Jianfeng Gao}, {and} \bibinfo{person}{Weizhu Chen}.} \bibinfo{year}{2020}\natexlab{}.
\newblock \showarticletitle{{Deberta: Decoding-enhanced bert with disentangled attention}}.
\newblock \bibinfo{journal}{\emph{arXiv preprint arXiv:2006.03654}} (\bibinfo{year}{2020}).
\newblock


\bibitem[Hu and Khan(2021)]%
        {hu2021uncertainty}
\bibfield{author}{\bibinfo{person}{Yibo Hu} {and} \bibinfo{person}{Latifur Khan}.} \bibinfo{year}{2021}\natexlab{}.
\newblock \showarticletitle{Uncertainty-aware reliable text classification}. In \bibinfo{booktitle}{\emph{Proceedings of the 27th ACM SIGKDD Conference on Knowledge Discovery \& Data Mining}}. \bibinfo{pages}{628--636}.
\newblock


\bibitem[Huang et~al\mbox{.}(2024)]%
        {calibrating_long}
\bibfield{author}{\bibinfo{person}{Yukun Huang}, \bibinfo{person}{Yixin Liu}, \bibinfo{person}{Raghuveer Thirukovalluru}, \bibinfo{person}{Arman Cohan}, {and} \bibinfo{person}{Bhuwan Dhingra}.} \bibinfo{year}{2024}\natexlab{}.
\newblock \showarticletitle{Calibrating Long-form Generations from Large Language Models}.
\newblock \bibinfo{journal}{\emph{arXiv preprint arXiv:2402.06544}} (\bibinfo{year}{2024}).
\newblock


\bibitem[Jha et~al\mbox{.}(2019)]%
        {jha2019attribution}
\bibfield{author}{\bibinfo{person}{Susmit Jha}, \bibinfo{person}{Sunny Raj}, \bibinfo{person}{Steven Fernandes}, \bibinfo{person}{Sumit~K Jha}, \bibinfo{person}{Somesh Jha}, \bibinfo{person}{Brian Jalaian}, \bibinfo{person}{Gunjan Verma}, {and} \bibinfo{person}{Ananthram Swami}.} \bibinfo{year}{2019}\natexlab{}.
\newblock \showarticletitle{Attribution-based confidence metric for deep neural networks}.
\newblock \bibinfo{journal}{\emph{Advances in Neural Information Processing Systems}}  \bibinfo{volume}{32} (\bibinfo{year}{2019}).
\newblock


\bibitem[Jiang et~al\mbox{.}(2023)]%
        {mistral}
\bibfield{author}{\bibinfo{person}{Albert~Q Jiang}, \bibinfo{person}{Alexandre Sablayrolles}, \bibinfo{person}{Arthur Mensch}, \bibinfo{person}{Chris Bamford}, \bibinfo{person}{Devendra~Singh Chaplot}, \bibinfo{person}{Diego de~las Casas}, \bibinfo{person}{Florian Bressand}, \bibinfo{person}{Gianna Lengyel}, \bibinfo{person}{Guillaume Lample}, \bibinfo{person}{Lucile Saulnier}, {et~al\mbox{.}}} \bibinfo{year}{2023}\natexlab{}.
\newblock \showarticletitle{Mistral 7B}.
\newblock \bibinfo{journal}{\emph{arXiv preprint arXiv:2310.06825}} (\bibinfo{year}{2023}).
\newblock


\bibitem[Jiang et~al\mbox{.}(2020)]%
        {log_likelihood}
\bibfield{author}{\bibinfo{person}{Zhengbao Jiang}, \bibinfo{person}{Frank~F Xu}, \bibinfo{person}{Jun Araki}, {and} \bibinfo{person}{Graham Neubig}.} \bibinfo{year}{2020}\natexlab{}.
\newblock \showarticletitle{How can we know what language models know?}
\newblock \bibinfo{journal}{\emph{Transactions of the Association for Computational Linguistics}}  \bibinfo{volume}{8} (\bibinfo{year}{2020}), \bibinfo{pages}{423--438}.
\newblock


\bibitem[Joshi et~al\mbox{.}(2017)]%
        {triviaqa}
\bibfield{author}{\bibinfo{person}{Mandar Joshi}, \bibinfo{person}{Eunsol Choi}, \bibinfo{person}{Daniel~S Weld}, {and} \bibinfo{person}{Luke Zettlemoyer}.} \bibinfo{year}{2017}\natexlab{}.
\newblock \showarticletitle{Triviaqa: A large scale distantly supervised challenge dataset for reading comprehension}.
\newblock \bibinfo{journal}{\emph{arXiv preprint arXiv:1705.03551}} (\bibinfo{year}{2017}).
\newblock


\bibitem[Kadavath et~al\mbox{.}(2022)]%
        {self_prob}
\bibfield{author}{\bibinfo{person}{Saurav Kadavath}, \bibinfo{person}{Tom Conerly}, \bibinfo{person}{Amanda Askell}, \bibinfo{person}{Tom Henighan}, \bibinfo{person}{Dawn Drain}, \bibinfo{person}{Ethan Perez}, \bibinfo{person}{Nicholas Schiefer}, \bibinfo{person}{Zac Hatfield-Dodds}, \bibinfo{person}{Nova DasSarma}, \bibinfo{person}{Eli Tran-Johnson}, {et~al\mbox{.}}} \bibinfo{year}{2022}\natexlab{}.
\newblock \showarticletitle{Language models (mostly) know what they know}.
\newblock \bibinfo{journal}{\emph{arXiv preprint arXiv:2207.05221}} (\bibinfo{year}{2022}).
\newblock


\bibitem[Kaur et~al\mbox{.}(2022)]%
        {idecode}
\bibfield{author}{\bibinfo{person}{Ramneet Kaur}, \bibinfo{person}{Susmit Jha}, \bibinfo{person}{Anirban Roy}, \bibinfo{person}{Sangdon Park}, \bibinfo{person}{Edgar Dobriban}, \bibinfo{person}{Oleg Sokolsky}, {and} \bibinfo{person}{Insup Lee}.} \bibinfo{year}{2022}\natexlab{}.
\newblock \showarticletitle{i{DECOD}e: In-distribution equivariance for conformal out-of-distribution detection}. In \bibinfo{booktitle}{\emph{Proceedings of the AAAI Conference on Artificial Intelligence}}, Vol.~\bibinfo{volume}{36}. \bibinfo{pages}{7104--7114}.
\newblock


\bibitem[Kaur et~al\mbox{.}(2023)]%
        {codit}
\bibfield{author}{\bibinfo{person}{Ramneet Kaur}, \bibinfo{person}{Kaustubh Sridhar}, \bibinfo{person}{Sangdon Park}, \bibinfo{person}{Yahan Yang}, \bibinfo{person}{Susmit Jha}, \bibinfo{person}{Anirban Roy}, \bibinfo{person}{Oleg Sokolsky}, {and} \bibinfo{person}{Insup Lee}.} \bibinfo{year}{2023}\natexlab{}.
\newblock \showarticletitle{CODiT: Conformal out-of-distribution Detection in time-series data for cyber-physical systems}. In \bibinfo{booktitle}{\emph{Proceedings of the ACM/IEEE 14th International Conference on Cyber-Physical Systems (with CPS-IoT Week 2023)}}. \bibinfo{pages}{120--131}.
\newblock


\bibitem[Kaur et~al\mbox{.}(2024)]%
        {kaur2024out}
\bibfield{author}{\bibinfo{person}{Ramneet Kaur}, \bibinfo{person}{Yahan Yang}, \bibinfo{person}{Oleg Sokolsky}, {and} \bibinfo{person}{Insup Lee}.} \bibinfo{year}{2024}\natexlab{}.
\newblock \showarticletitle{Out-of-Distribution Detection in Dependent Data for Cyber-Physical Systems with Conformal Guarantees}.
\newblock \bibinfo{journal}{\emph{ACM Transactions on Cyber-Physical Systems}} (\bibinfo{year}{2024}).
\newblock


\bibitem[Kuhn et~al\mbox{.}(2023)]%
        {kuhn}
\bibfield{author}{\bibinfo{person}{Lorenz Kuhn}, \bibinfo{person}{Yarin Gal}, {and} \bibinfo{person}{Sebastian Farquhar}.} \bibinfo{year}{2023}\natexlab{}.
\newblock \showarticletitle{Semantic uncertainty: Linguistic invariances for uncertainty estimation in natural language generation}.
\newblock \bibinfo{journal}{\emph{arXiv preprint arXiv:2302.09664}} (\bibinfo{year}{2023}).
\newblock


\bibitem[Lin(2004)]%
        {rouge}
\bibfield{author}{\bibinfo{person}{Chin-Yew Lin}.} \bibinfo{year}{2004}\natexlab{}.
\newblock \showarticletitle{Rouge: A package for automatic evaluation of summaries}. In \bibinfo{booktitle}{\emph{Text summarization branches out}}. \bibinfo{pages}{74--81}.
\newblock


\bibitem[Lin et~al\mbox{.}(2023)]%
        {lin}
\bibfield{author}{\bibinfo{person}{Zhen Lin}, \bibinfo{person}{Shubhendu Trivedi}, {and} \bibinfo{person}{Jimeng Sun}.} \bibinfo{year}{2023}\natexlab{}.
\newblock \showarticletitle{Generating with confidence: Uncertainty quantification for black-box large language models}.
\newblock \bibinfo{journal}{\emph{arXiv preprint arXiv:2305.19187}} (\bibinfo{year}{2023}).
\newblock


\bibitem[Magesh et~al\mbox{.}(2023)]%
        {Magesh2023PrincipledOD}
\bibfield{author}{\bibinfo{person}{Akshayaa Magesh}, \bibinfo{person}{Venugopal~V. Veeravalli}, \bibinfo{person}{Anirban Roy}, {and} \bibinfo{person}{Susmit Jha}.} \bibinfo{year}{2023}\natexlab{}.
\newblock \showarticletitle{Principled OOD Detection via Multiple Testing}.
\newblock \bibinfo{journal}{\emph{2023 IEEE International Symposium on Information Theory (ISIT)}} (\bibinfo{year}{2023}), \bibinfo{pages}{1026--1031}.
\newblock
\urldef\tempurl%
\url{https://api.semanticscholar.org/CorpusID:261081329}
\showURL{%
\tempurl}


\bibitem[Mielke et~al\mbox{.}(2022)]%
        {reducing_overconf_cali}
\bibfield{author}{\bibinfo{person}{Sabrina~J Mielke}, \bibinfo{person}{Arthur Szlam}, \bibinfo{person}{Emily Dinan}, {and} \bibinfo{person}{Y-Lan Boureau}.} \bibinfo{year}{2022}\natexlab{}.
\newblock \showarticletitle{Reducing conversational agents’ overconfidence through linguistic calibration}.
\newblock \bibinfo{journal}{\emph{Transactions of the Association for Computational Linguistics}}  \bibinfo{volume}{10} (\bibinfo{year}{2022}), \bibinfo{pages}{857--872}.
\newblock


\bibitem[Nadeem et~al\mbox{.}(2009)]%
        {auarc}
\bibfield{author}{\bibinfo{person}{Malik Sajjad~Ahmed Nadeem}, \bibinfo{person}{Jean-Daniel Zucker}, {and} \bibinfo{person}{Blaise Hanczar}.} \bibinfo{year}{2009}\natexlab{}.
\newblock \showarticletitle{Accuracy-rejection curves (ARCs) for comparing classification methods with a reject option}. In \bibinfo{booktitle}{\emph{Machine Learning in Systems Biology}}. PMLR, \bibinfo{pages}{65--81}.
\newblock


\bibitem[Quach et~al\mbox{.}(2023)]%
        {conformal_lang_modeling}
\bibfield{author}{\bibinfo{person}{Victor Quach}, \bibinfo{person}{Adam Fisch}, \bibinfo{person}{Tal Schuster}, \bibinfo{person}{Adam Yala}, \bibinfo{person}{Jae~Ho Sohn}, \bibinfo{person}{Tommi~S Jaakkola}, {and} \bibinfo{person}{Regina Barzilay}.} \bibinfo{year}{2023}\natexlab{}.
\newblock \showarticletitle{Conformal language modeling}.
\newblock \bibinfo{journal}{\emph{arXiv preprint arXiv:2306.10193}} (\bibinfo{year}{2023}).
\newblock


\bibitem[Reddy et~al\mbox{.}(2019)]%
        {coqa}
\bibfield{author}{\bibinfo{person}{Siva Reddy}, \bibinfo{person}{Danqi Chen}, {and} \bibinfo{person}{Christopher~D Manning}.} \bibinfo{year}{2019}\natexlab{}.
\newblock \showarticletitle{Coqa: A conversational question answering challenge}.
\newblock \bibinfo{journal}{\emph{Transactions of the Association for Computational Linguistics}}  \bibinfo{volume}{7} (\bibinfo{year}{2019}), \bibinfo{pages}{249--266}.
\newblock


\bibitem[Touvron et~al\mbox{.}(2023)]%
        {llama-2}
\bibfield{author}{\bibinfo{person}{Hugo Touvron}, \bibinfo{person}{Louis Martin}, \bibinfo{person}{Kevin Stone}, \bibinfo{person}{Peter Albert}, \bibinfo{person}{Amjad Almahairi}, \bibinfo{person}{Yasmine Babaei}, \bibinfo{person}{Nikolay Bashlykov}, \bibinfo{person}{Soumya Batra}, \bibinfo{person}{Prajjwal Bhargava}, \bibinfo{person}{Shruti Bhosale}, {et~al\mbox{.}}} \bibinfo{year}{2023}\natexlab{}.
\newblock \showarticletitle{Llama 2: Open foundation and fine-tuned chat models}.
\newblock \bibinfo{journal}{\emph{arXiv preprint arXiv:2307.09288}} (\bibinfo{year}{2023}).
\newblock


\bibitem[Tuncer and Schulz(2016)]%
        {tuncer2016sequential}
\bibfield{author}{\bibinfo{person}{Mehmet Ali~{\c{C}}a{\u{g}}r{\i} Tuncer} {and} \bibinfo{person}{Dirk Schulz}.} \bibinfo{year}{2016}\natexlab{}.
\newblock \showarticletitle{Sequential distance dependent chinese restaurant processes for motion segmentation of 3d lidar data}. In \bibinfo{booktitle}{\emph{2016 19th International Conference on Information Fusion (FUSION)}}. IEEE, \bibinfo{pages}{758--765}.
\newblock


\bibitem[Vovk et~al\mbox{.}(2005)]%
        {icp}
\bibfield{author}{\bibinfo{person}{Vladimir Vovk}, \bibinfo{person}{Alex Gammerman}, {and} \bibinfo{person}{Glenn Shafer}.} \bibinfo{year}{2005}\natexlab{}.
\newblock \bibinfo{booktitle}{\emph{Algorithmic learning in a random world}}.
\newblock \bibinfo{publisher}{Springer Science \& Business Media}.
\newblock


\bibitem[Wang et~al\mbox{.}(2022)]%
        {wang2022self}
\bibfield{author}{\bibinfo{person}{Xuezhi Wang}, \bibinfo{person}{Jason Wei}, \bibinfo{person}{Dale Schuurmans}, \bibinfo{person}{Quoc Le}, \bibinfo{person}{Ed Chi}, \bibinfo{person}{Sharan Narang}, \bibinfo{person}{Aakanksha Chowdhery}, {and} \bibinfo{person}{Denny Zhou}.} \bibinfo{year}{2022}\natexlab{}.
\newblock \showarticletitle{Self-consistency improves chain of thought reasoning in language models}.
\newblock \bibinfo{journal}{\emph{arXiv preprint arXiv:2203.11171}} (\bibinfo{year}{2022}).
\newblock


\bibitem[Xiao and Wang(2019)]%
        {xiao2019quantifying}
\bibfield{author}{\bibinfo{person}{Yijun Xiao} {and} \bibinfo{person}{William~Yang Wang}.} \bibinfo{year}{2019}\natexlab{}.
\newblock \showarticletitle{Quantifying uncertainties in natural language processing tasks}. In \bibinfo{booktitle}{\emph{Proceedings of the AAAI conference on artificial intelligence}}, Vol.~\bibinfo{volume}{33}. \bibinfo{pages}{7322--7329}.
\newblock


\bibitem[Yang et~al\mbox{.}(2024)]%
        {yang2024memory}
\bibfield{author}{\bibinfo{person}{Yahan Yang}, \bibinfo{person}{Ramneet Kaur}, \bibinfo{person}{Souradeep Dutta}, {and} \bibinfo{person}{Insup Lee}.} \bibinfo{year}{2024}\natexlab{}.
\newblock \showarticletitle{Memory-based Distribution Shift Detection for Learning Enabled Cyber-Physical Systems with Statistical Guarantees}.
\newblock \bibinfo{journal}{\emph{ACM Transactions on Cyber-Physical Systems}} \bibinfo{volume}{8}, \bibinfo{number}{2} (\bibinfo{year}{2024}), \bibinfo{pages}{1--28}.
\newblock


\bibitem[Ye et~al\mbox{.}(2024)]%
        {UQ_benchmark}
\bibfield{author}{\bibinfo{person}{Fanghua Ye}, \bibinfo{person}{Mingming Yang}, \bibinfo{person}{Jianhui Pang}, \bibinfo{person}{Longyue Wang}, \bibinfo{person}{Derek~F Wong}, \bibinfo{person}{Emine Yilmaz}, \bibinfo{person}{Shuming Shi}, {and} \bibinfo{person}{Zhaopeng Tu}.} \bibinfo{year}{2024}\natexlab{}.
\newblock \showarticletitle{Benchmarking LLMs via Uncertainty Quantification}.
\newblock \bibinfo{journal}{\emph{arXiv preprint arXiv:2401.12794}} (\bibinfo{year}{2024}).
\newblock


\end{thebibliography}
\bibliographystyle{ACM-Reference-Format}

\newpage
\newpage
\appendix
\newpage 

\newpage 
\section{Appendix}
\label{sec:appendix}

\subsection{Qualitative comparison of Clustering Approaches}
\label{app:qualcluster}
Here, we provide an example from both COQA and TriviaQA datasets to analyse how our new clustering approach compares to the original approach by \citet{kuhn}. We look at the quality of clusters formed by both approaches.

\subsubsection{An example from COQA Dataset}
\textbf{Story}: CHAPTER XXXIV Arthur remained at the gate while Ruth climbed Maria's front steps. She heard the rapid click of the type-writer, and when Martin let her in, found him on the last page of a manuscript. She had come to make certain whether or not he would be at their table for Thanksgiving dinner; but before she could broach the subject Martin plunged into the one with which he was full. ``Here, let me read you this,'' he cried, separating the carbon copies and running the pages of manuscript into shape. ``It's my latest, and different from anything I've done. It is so altogether different that I am almost afraid of it, and yet I've a sneaking idea it is good. You be judge. It's an Hawaiian story. I've called it `Wiki-wiki'.'' His face was bright with the creative glow, though she shivered in the cold room and had been struck by the coldness of his hands at greeting. She listened closely while he read, and though he from time to time had seen only disapprobation in her face, at the close he asked:- ``Frankly, what do you think of it?'' ``I--I don't know,'' she, answered. ``Will it--do you think it will sell?'' ``I'm afraid not,'' was the confession. ``It's too strong for the magazines. But it's true, on my word it's true.'' ``But why do you persist in writing such things when you know they won't sell?'' she went on inexorably. ``The reason for your writing is to make a living, isn't it?''
\\
\textbf{Question}: `Did he answer her?'
\\
\textbf{Answer}: `No'
\\
\textbf{Generated Responses from Llama-13b}: [`Yes', `Yes', ``He didn't'', `He did, only not directly', `No', `No', `No', `He asked her what she thought', `He told her his latest story', `Yes', `No', `A sneaking yes', `He ran the manuscript up to Miss Lawton', `No', `In the affirmative', `Yes', `No', `No', `Yes', `Yes']

\begin{figure*}[!h]
    \centering
\includegraphics[width=1\columnwidth]{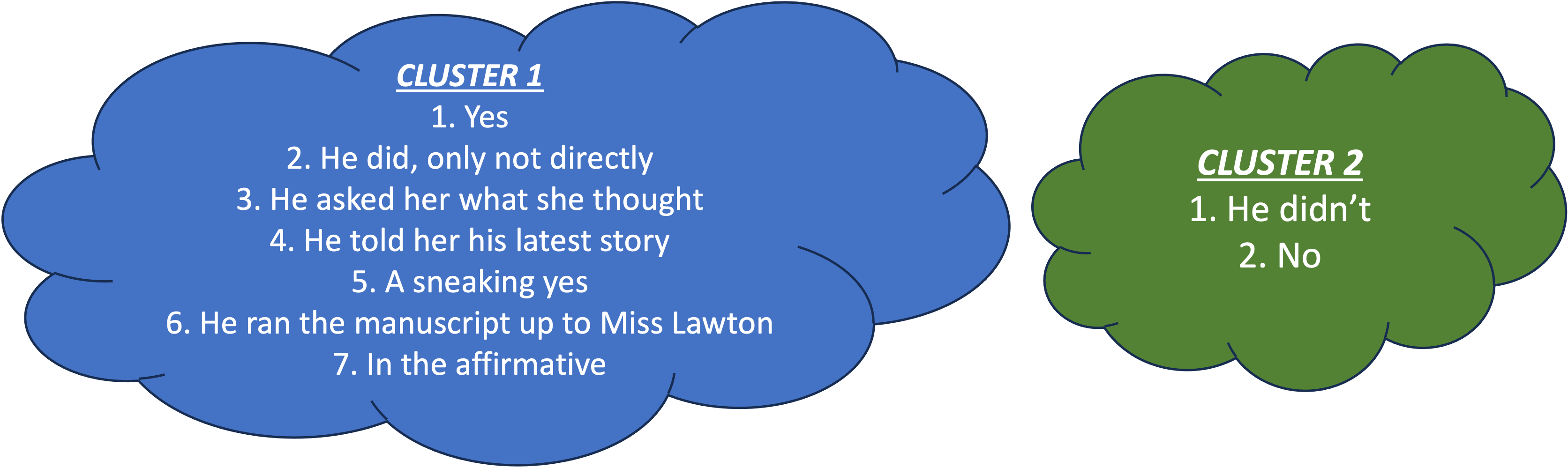}
    \caption{Clusters generated by ~\citet{kuhn}'s approach on COQA example.}
    \label{fig:kuhn_clusters_coqa}
\end{figure*}
\begin{figure*}[!h]
    \centering
\includegraphics[width=1\columnwidth]{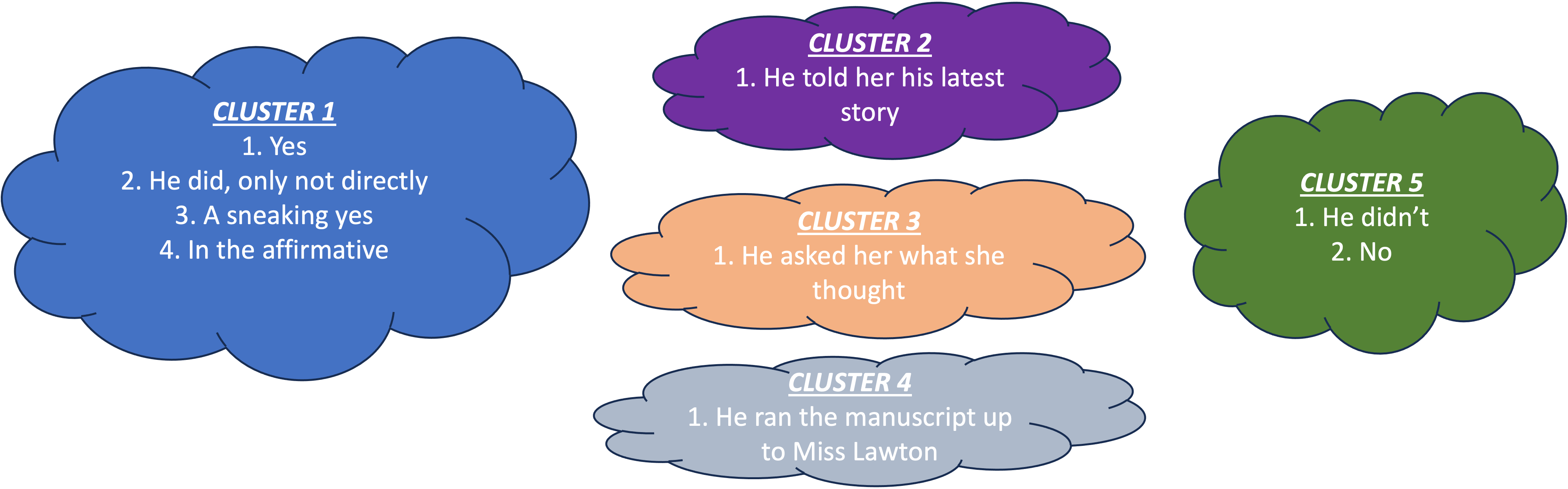}
    \caption{Clusters generated by our approach (Alg.~\ref{alg:clustering}) on COQA example.}
    \label{fig:our_clusters_coqa}
\end{figure*}

\textbf{Results}: Figures~\ref{fig:kuhn_clusters_coqa}, and~\ref{fig:our_clusters_coqa} show the clusters formed by~\citet{kuhn}, and our approach respectively.  For brevity, we include only unique responses in a cluster. As it can be seen,~\citet{kuhn} approach puts semantically different responses in the same cluster (responses 3, 4, and 6 in cluster 1 for `Yes'), whereas ours separate them out in different clusters.

\subsubsection{An example from TriviaQA Dataset}
\textbf{Question}: What is `The Old Lady of Threadneedle Street'?
\\
\textbf{Answer}: Bank of England
\\
\textbf{Generated Responses from Llama-13b}: [`Bank of England', `Bank of England', `Bank of England', `The Bank of England', `A nickname; what was it really?', `Bank of England', `The Bank of England', `The Bank of England', `The Bank of England', `The Bank of England', `The Bank of England', `Bank of England', `The Bank of England', `Bank of England', `The Bank of England', `Bank of England', `Bank of England', `Bank of England', `The Bank of England', `Bank Of England']
\\
\textbf{Results}: Figures~\ref{fig:kuhn_clusters_coqa}, and~\ref{fig:our_clusters_coqa} show the clusters formed by~\citet{kuhn}, and our approach respectively.  Again for brevity, we include only unique responses in a cluster. As it can be seen,~\citet{kuhn} approach puts semantically different responses in the same cluster (response 3 in cluster 1 for `Bank of England'), whereas ours separate them out in different clusters.

We observed similar results on other stories from COQA, and questions from TriviaQA datasets.

\begin{figure*}[!h]
    \centering
\includegraphics[width=0.5\columnwidth]{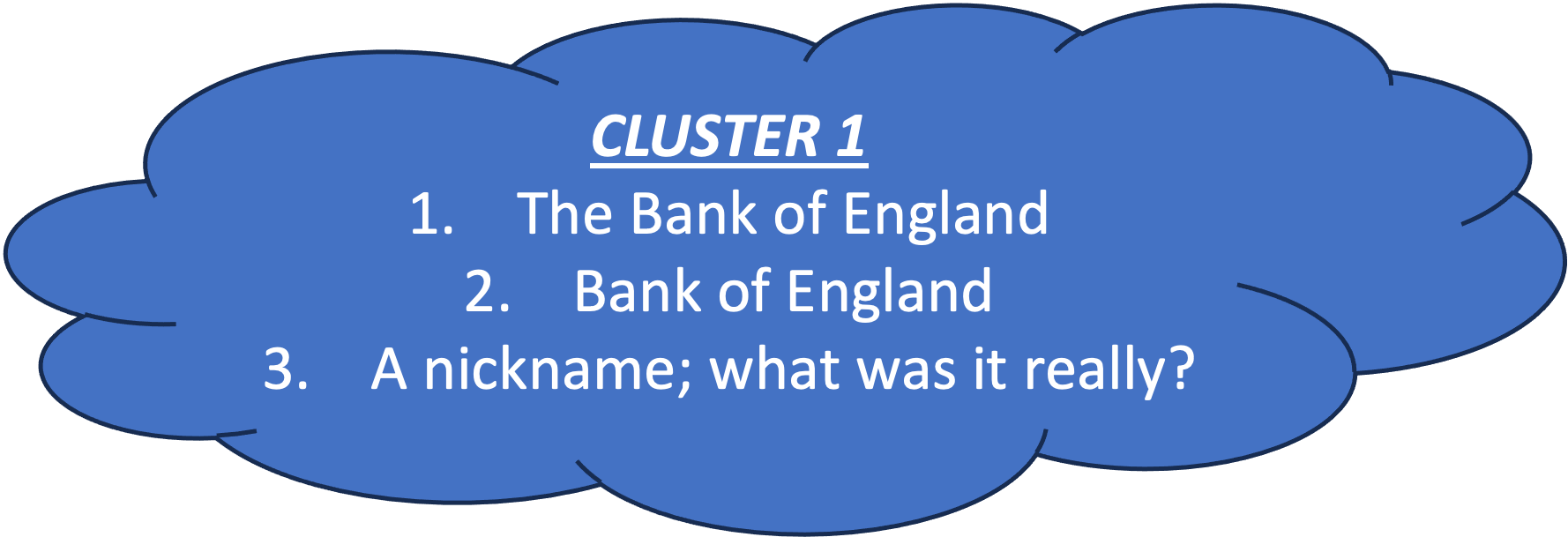}
    \caption{Clusters generated by ~\citet{kuhn}'s approach on TriviaQA example.}
    \label{fig:kuhn_clusters_triviaiqa}
\end{figure*}

\begin{figure*}[!h]
    \centering
\includegraphics[width=1\columnwidth]{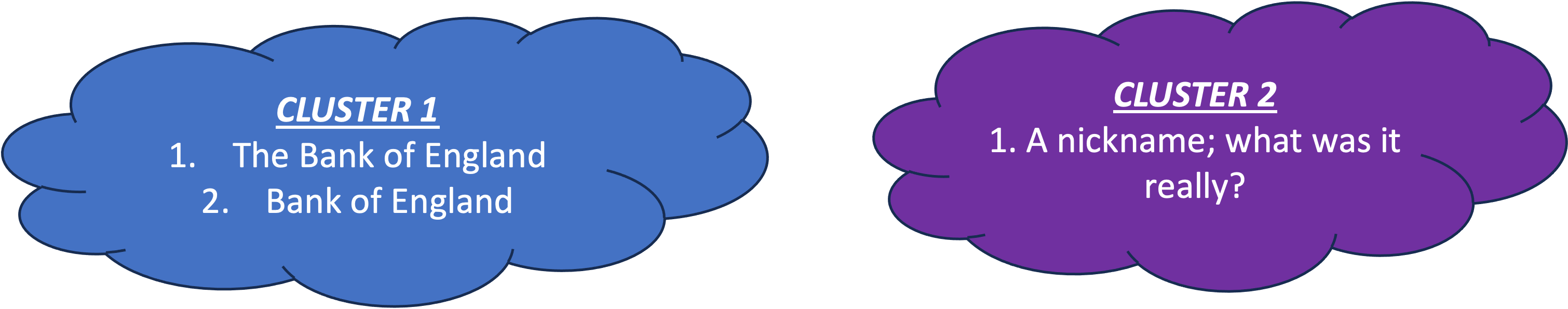}
    \caption{Clusters generated by our approach (Alg.~\ref{alg:clustering}) on TriviaQA example.}
    \label{fig:our_clusters_triviaqa}
\end{figure*}

\subsection{All UQ results}
\label{app_sec_all_uq_results}
Here, we include all results on UQ performance from Section~\ref{sec:uq_perf}: comparison with additional baselines on AUARC, \newMetricName\, and AUROC evaluation metrics. In addition to ~\citet{kuhn}'s Sem. Ent. (Unnorm/Norm), and~\citep{lin}'s EigV results reported in the main paper, we include  ``Numset'', ``LexiSim'', and ``SelfProb'' baselines here. Numset uses the number of semantic sets (or clusters) as the UQ metric, and has been previously used in~\citep{lin} as one of the baselines. Higher the numset, more uncertain is the LLM on the input query. LexiSim uses the average of RougeL distance between every pair of responses for UQ. Here, higher the Lexisim, lower is the uncertainty.
Again, Lexisim has been used as a baseline by~\citet{kuhn}, and~\citet{lin}. SelfProb~\citep{self_prob} estimates if the probability of a model’s response is correct by asking the model itself, and use that as the UQ metric. We follow the same prompt format as~\citet{lin} for asking the model about the probability, and report average over all responses. Here, higher the SelfProb, lower is the uncertainty.

Tables~\ref{tab:all_auarc_results_coqa},~\ref{tab:all_auroc_results_coqa}, and~\ref{tab:all_rej_acc_results_coqa} are AUARC, AUROC, and \newMetricName\ are results on COQA. And  Tables~\ref{tab:all_auarc_results_triviaqa},~\ref{tab:all_auroc_results_triviaqa}, and~\ref{tab:all_rej_acc_results_triviaqa} are AUARC, AUROC, and \newMetricName\ are results on TriviaQA. Again, we report all results from the three GT evaluation methods: GPT-4, RougeL, and Deberta. We achieve either the best or second best in all but two test cases.


\begin{table*}[!h]
    \centering
    \setlength{\tabcolsep}{2pt}
    \resizebox{1\columnwidth}{!}{
    \begin{tabular}{c|c|c|c|c|c|c|c|c}
    \hline
        \textbf{Model} & \textbf{GT} & \textbf{Model Acc} & \textbf{Sem. Ent.} & \textbf{NumSet} & \textbf{LexiSim} & \textbf{SelfProb} & \textbf{EigV} & \textbf{Ours} \\
        & & & Unnorm/Norm & & & & & Unnorm/Norm
        \\
        \hline
        Llama-13b & GPT-4 & 73.22 & 85.81/86.44 & 79.78 & 86.14 & 75.50 & \textbf{88.03} & 86.35/\underline{87.47}\\
        Mistral-7b & GPT-4 & 73.38 &  81.91/82.68 & 75.63 & 81.73 & \textbf{85.14} &  82.82 & 82.22/\underline{82.95}\\
        \hline
        Mean & GPT-4 & 73.30 & 83.86/84.56 & 77.71 & 83.94 & 80.32 & \textbf{85.43} & 84.29/\underline{85.21}\\
        \hline
        Llama-13b & RougeL & 72.75 &  86.03/87.05 & 77.79 & \underline{88.17} & 73.35 & 87.92 & 86.84/\textbf{88.34}\\
        Mistral-7b & RougeL & 44.74 & \underline{64.37}/62.93  & 46.99 & 59.61 & 52.64 & 63.43 & \textbf{64.60}/63.48 \\
        \hline
        Mean & RougeL & 58.75 & 75.20/74.99 & 62.39 & 73.89 & 63.00 & 75.65  & \underline{75.72}/\textbf{75.91}  \\
        \hline
        Llama-13b & Deberta & 63.74 & 80.21/79.48 & 69.36  & 79.02 & 65.23 & \textbf{82.68} & 81.04/\underline{81.37} \\
        Mistral-7b & Deberta & 11.23 & \underline{23.56}/20.71 & 11.63 & 16.70 & 12.21 & 20.88 & \textbf{23.53}/21.05\\
        \hline
        Mean & Deberta & 37.49 & \underline{51.89}/50.10& 40.50 & 47.86 & 38.72 & 51.78 & \textbf{52.29}/51.21 \\
        \hline
    \end{tabular}
    }
    \caption{AUARC ($\uparrow$) results in comparison to all baselines on \textbf{COQA}. Best results are in bold and second best are underlined.}
    \label{tab:all_auarc_results_coqa}
\end{table*}

\begin{table*}[!h]
    \centering
    \setlength{\tabcolsep}{2pt}
    \resizebox{1\columnwidth}{!}{
    \begin{tabular}{c|c|c|c|c|c|c|c|c}
    \hline
        \textbf{Model} & \textbf{GT} & \textbf{Model Acc} & \textbf{Sem. Ent.} & \textbf{NumSet} & \textbf{LexiSim} & \textbf{SelfProb} & \textbf{EigV} & \textbf{Ours} \\
        & & & Unnorm/Norm & & & &  & Unnorm/Norm
        \\
        \hline
        Llama-13b & GPT-4 & 67.03 & 88.13/87.94  & 83.84 & 84.52 & 73.09 & \textbf{88.84} & 88.33/\underline{88.54} \\
        Mistral-7b & GPT-4 & 60.68 & 80.99/81.40 & 74.72 & 76.65 & \textbf{84.46} & \underline{82.03} & 81.23/\underline{82.03} \\
        \hline
        Mean & GPT-4 & 63.86 & 84.56/84.67 & 79.28 & 80.59 & 78.78 & \textbf{85.44} & 84.78/\underline{85.29} \\
        \hline
        Llama-13b & RougeL & 64.60 & 85.62/85.19 & 79.75  & 84.01 & 70.34 & 85.76 & \underline{85.86}/\textbf{85.87} \\
        Mistral-7b & RougeL & 42.33 & \underline{70.18}/68.13 & 54.53 & 61.72 & 62.03 & 69.41 & \textbf{70.26}/68.81 \\
        \hline
        Mean & RougeL & 53.47 & \underline{77.90}/76.66 & 67.14 & 72.87 & 66.19 & 77.59 & \textbf{78.06}/77.34 \\
        \hline
        Llama-13b & Deberta & 63.33 & 84.92/84.34  & 79.11 & 80.01 & 68.04 & \textbf{85.60} & \underline{85.23}/85.13 \\
        Mistral-7b & Deberta & 33.92 & \underline{62.29}/59.53 & 44.88 & 51.80 & 50.33 & 60.39 & \textbf{62.37}/60.16 \\
        \hline
        Mean & Deberta & 48.63 & \underline{73.61}/71.94 & 62.00 & 65.91 & 59.19 & 73.00 & \textbf{73.80}/72.65 \\
        \hline
    \end{tabular}
    }
    
    \caption{AUARC ($\uparrow$) results in comparison to all baselines on \textbf{TriviaQA}. Best results are in bold and second best are underlined.}
    \label{tab:all_auarc_results_triviaqa}
\end{table*}

\begin{table*}[!h]
    \centering
    \setlength{\tabcolsep}{2pt}
    \resizebox{1\columnwidth}{!}{
    \begin{tabular}{c|c|c|c|c|c|c|c|c}
    \hline
        \textbf{Model} & \textbf{GT} & \textbf{Model Acc} & \textbf{Sem. Ent.} & \textbf{NumSet} & \textbf{LexiSim} & \textbf{SelfProb} & \textbf{EigV} & \textbf{Ours} \\
        & & & Unnorm/Norm & & & &  & Unnorm/Norm
        \\
        \hline
        Llama-13b & GPT-4 & 73.22 & 85.19/88.69 & 73.06 & 82.63 & 53.74 & \textbf{92.83} & 87.87/\underline{91.90} \\
        Mistral-7b & GPT-4 & 73.38 &  79.91/81.99 & 58.00 & 76.57 & 79.46 &  \underline{82.77} & 81.48/\textbf{84.07}\\
        \hline
        Mean & GPT-4 & 73.3  & 82.55/85.34 & 65.53 & 79.6  & 66.6  & \underline{87.80}  & 84.68/\textbf{87.99}  \\
        \hline
        Llama-13b & RougeL & 72.75 &  80.87/86.48 & 68.24 & \textbf{92.33} & 48.40 & 88.45 & 83.90/\underline{90.28}\\
        Mistral-7b & RougeL & 44.74 & 83.52/83.97  & 55.43 & 83.15 & 70.62 & \textbf{86.84} & 84.63/\underline{85.69} \\
        \hline
        Mean & RougeL & 58.75 & 82.20/85.23 & 61.84 & \underline{87.74} & 59.51 & 87.65 & 84.27/\textbf{87.99} \\
        \hline
        Llama-13b & Deberta & 63.74 & 85.59/88.80 & 69.55  & 87.45 & 48.92 & \textbf{93.09} & 88.38/\underline{92.02} \\
        Mistral-7b & Deberta & 11.23 & \underline{93.98}/91.22 & 54.62 & 91.57 & 59.48 & 93.85 & \textbf{94.25}/92.07\\
        \hline
        Mean & Deberta & 37.485& 89.79/90.01 & 62.09 & 89.51 & 54.2  & \textbf{93.47} & 91.32/\underline{92.05} \\
    \hline
    \end{tabular}
    }
    
    \caption{AUROC ($\uparrow$) results in comparison to all baselines on \textbf{COQA}. Best results are in bold and second best are underlined.}
    \label{tab:all_auroc_results_coqa}
\end{table*}

\begin{table*}[!h]
    \centering
    \setlength{\tabcolsep}{2pt}
    \resizebox{1\columnwidth}{!}{
    \begin{tabular}{c|c|c|c|c|c|c|c|c}
    \hline
        \textbf{Model} & \textbf{GT} & \textbf{Model Acc} & \textbf{Sem. Ent.} & \textbf{NumSet} & \textbf{LexiSim} & \textbf{SelfProb} & \textbf{EigV} & \textbf{Ours} \\
        & & & Unnorm/Norm & & & &  & Unnorm/Norm
        \\
        \hline
        Llama-13b & GPT-4 & 67.03 & 94.74/96.85  & 92.04 & 86.64 & 59.38 & \textbf{97.48} & 95.29/\underline{97.39} \\
        Mistral-7b & GPT-4 & 60.68 & 90.58/93.59 & 80.93 & 81.00 & \textbf{94.75} & 93.66 & 91.48/\underline{94.24} \\
        \hline
        Mean & GPT-4 & 63.86 & 92.66/95.22 & 86.49 & 83.82 & 77.07 & \underline{95.57} & 93.39/\textbf{95.82} \\
        \hline
        Llama-13b & RougeL & 64.60 & 92.63/94.50 & 88.59  & 97.01 & 59.84 & \textbf{95.25} & 93.27/\underline{95.23}\\
        Mistral-7b & RougeL & 42.33 & 95.26/94.80 & 74.87 & 94.66 & 84.67 & \textbf{96.34} & \underline{95.61}/95.36 \\
        \hline
        Mean & RougeL & 53.47 & 93.95/94.65 & 81.73 & \textbf{95.84} & 72.26 & \underline{95.80} & 94.44/95.30 \\
        \hline
        Llama-13b & Deberta & 63.33 & 92.61/94.34  & 87.73 & 86.58 & 55.96 & \textbf{96.55} & 93.49/\underline{95.30} \\
        Mistral-7b & Deberta & 33.92 & 96.55/94.90 & 73.84 & 93.06 & 82.01 & \underline{96.64} & \textbf{96.73}/95.30 \\
        \hline
        Mean & Deberta & 48.63 & 94.58/94.62 & 80.79 & 89.82 & 68.99 & \textbf{96.60} & 95.11/\underline{95.30}  \\
        \hline
    \end{tabular}
    }
    
    \caption{AUROC ($\uparrow$) results in comparison to all baselines on \textbf{TriviaQA}. Best results are in bold and second best are underlined.}
    \label{tab:all_auroc_results_triviaqa}
\end{table*}

\begin{table*}[!h]
    \centering
    \setlength{\tabcolsep}{2pt}
    \resizebox{1\columnwidth}{!}{
    \begin{tabular}{c|c|c|c|c|c|c|c|c}
    \hline
        \textbf{Model} & \textbf{GT} & \textbf{Model Acc} & \textbf{Sem. Ent.} & \textbf{NumSet} & \textbf{LexiSim} & \textbf{SelfProb} & \textbf{EigV} & \textbf{Ours} \\
        & & & Unnorm/Norm & & & &  & Unnorm/Norm
        \\
        \hline
        Llama-13b & GPT-4 & 73.22 & 58.97/56.90 & 63.45 & 61.35  & 72.29 & \textbf{54.63} &  58.42/\underline{55.32} \\
        Mistral-7b & GPT-4 & 73.38 & 63.06/62.02 & 67.19 & 62.21 & \underline{61.18} & \textbf{59.83} & 62.77/61.41 \\
        \hline
        Mean & GPT-4 & 73.30  & 61.02/59.46 & 65.32 & 61.78 & 66.74 & \textbf{57.23} & 60.60/\underline{58.37} \\
        \hline
        Llama-13b & RougeL & 72.75 & 56.75/55.53 & 65.30 & 58.51  & 73.12 & \underline{53.78} & 55.78/\textbf{52.65} \\
        Mistral-7b & RougeL & 44.74 & 27.62/29.65 & 40.13 & 34.11 & 33.56 & \textbf{27.12} & \underline{27.37}/28.26 \\
        \hline
        Mean & RougeL & 58.75 & 42.19/42.59 & 52.72& 46.31 & 53.34 & \textbf{40.45} & 41.58/\underline{40.46}\\
        \hline
        Llama-13b & Deberta &  63.74 & 46.07/46.91 & 55.50 &  53.07 & 63.32 & \textbf{42.04} & 45.07/\underline{43.56} \\
        Mistral-7b & Deberta & 11.23 & \underline{3.84}/5.70 & 9.84 & 11.83 & 9.45 & 4.13 & \textbf{3.82}/5.00\\
        \hline
        Mean & Deberta & 37.49 & 24.96/26.31 & 32.67 & 32.45 & 36.39& \textbf{23.09} & \underline{24.45}/24.28\\
        \hline
    \end{tabular}
    }
    
    \caption{\newMetricName\ ($\downarrow$) results in comparison to all baselines on \textbf{COQA}. Best results are in bold and second best are underlined.}
    \label{tab:all_rej_acc_results_coqa}
\end{table*}

\begin{table*}[!h]
    \centering
    \setlength{\tabcolsep}{2pt}
    \resizebox{1\columnwidth}{!}{
    \begin{tabular}{c|c|c|c|c|c|c|c|c}
    \hline
        \textbf{Model} & \textbf{GT} & \textbf{Model Acc} & \textbf{Sem. Ent.} & \textbf{NumSet} & \textbf{LexiSim} & \textbf{SelfProb} & \textbf{EigV} & \textbf{Ours} \\
        & & & Unnorm/Norm & & & &  & Unnorm/Norm
        \\
        \hline
        Llama-13b & GPT-4 & 67.03 & 40.09/40.27 & 43.03  & 54.28 & 60.31 & \underline{39.42} & 39.92/\textbf{39.38} \\
        Mistral-7b & GPT-4 & 60.68 & 35.57/35.04 & 40.02 & 47.45 & 33.71 & \textbf{33.19} & 35.13/\underline{33.29}\\
        \hline
        Mean & GPT-4 & 63.86& 37.83/37.66 & 41.525& 50.87& 47.01 & \textbf{36.31} & 37.53/\underline{36.34} \\
        \hline
        Llama-13b & RougeL & 64.60 & 39.12/39.39 & 42.74 & 50.56 & 58.41 & 38.93 &  \underline{38.81}/\textbf{38.35} \\
        Mistral-7b & RougeL & 42.33& 17.15/19.56  & 25.09 & 32.12 & 21.46 & \underline{17.06} & \textbf{16.95}/18.11 \\
        \hline
        Mean & RougeL & 53.47& 28.14/29.48& 33.915& 41.34 & 39.94& \underline{28.00} & \textbf{27.88}/28.23 \\
        \hline
        Llama-13b & Deberta & 63.33 & 37.23/37.94 & 40.09 & 53.01 & 58.23 & \textbf{36.70} & \underline{36.88}/36.84 \\
        Mistral-7b & Deberta & 33.92 & \underline{11.00}/13.54 & 18.51 & 27.93 & 15.97 &  11.35 & \textbf{10.89}/12.45\\
        \hline
        Mean & Deberta & 48.63 & 24.12/25.74 & 29.3  & 40.47 & 37.1  & \underline{24.03} & \textbf{23.89}/24.65 \\
        \hline
    \end{tabular}
    }
    
    \caption{\newMetricName\ ($\downarrow$) results in comparison to all baselines on \textbf{TriviaQA}. Best results are in bold and second best are underlined.}
    \label{tab:all_rej_acc_results_triviaqa}
\end{table*}

\newpage

\end{document}